\documentclass[journal]{IEEEtran}

\usepackage{ifpdf}
\usepackage{cite}
\usepackage{graphicx}
\usepackage{textcomp}
\usepackage{amsmath}
\usepackage{color}
\usepackage{url}
\usepackage{fancyhdr}
\usepackage{subfig}
\usepackage{makecell}
\usepackage{amssymb}
\usepackage{amsmath}
\usepackage{algorithmic}
\usepackage{array}
\usepackage{longtable}
\usepackage{booktabs}
\usepackage{supertabular}
\begin{document}

\title{Deep Gated Recurrent and Convolutional Network Hybrid Model for Univariate Time Series Classification
}

\author{Nelly~Elsayed,~\IEEEmembership{Student Member,~IEEE,}
        Anthony~S.~Maida
        and~Magdy~Bayoumi,~\IEEEmembership{Life~Fellow,~IEEE}% <-this % stops a space
}


\maketitle

\begin{abstract}
Hybrid LSTM-fully convolutional networks (LSTM-FCN) for time series classification have produced state-of-the-art classification results
on univariate time series. We show that replacing the LSTM with a gated recurrent unit (GRU) to create a GRU-fully convolutional network hybrid model (GRU-FCN) can offer even better performance on many time series datasets. The proposed GRU-FCN model outperforms state-of-the-art classification performance in many univariate time series datasets without additional supporting algorithms requirement.
Furthermore, since the GRU uses a simpler architecture than the LSTM,
it has fewer training parameters, less training time, smaller memory storage requirement, and a simpler hardware implementation, compared to the LSTM-based models.
\end{abstract}


\IEEEpeerreviewmaketitle

\section{Introduction}
A time series (TS) is a sequence of data points obtained at successive equally-spaced time points, ordinarily in a uniform interval time domain~\cite{hamilton1994time}. 
TSs are used in several research and industrial fields where temporal analysis measurements are involved such as in signal processing~\cite{sohn2001damage}, pattern recognition~\cite{gul2009statistical}, mathematics~\cite{hamilton1994time}, psychological and physiological signals analysis~\cite{wang2017time, karim2018lstm}, earthquake prediction~\cite{amei2012time}, weather readings~\cite{rotton1985air}, and statistics~\cite{hamilton1994time}. 
There are two types of time series: univariate and multivariate. 
%The multivariate time series is more intricate and complex than the univariate time series because it has multiple varying variable dependencies over a period of time but the univariate has only one varying variable over time~\cite{hamilton1994time}. 
In this paper, we study the univariate time series classification.

There are many approaches to time series classification. 
The distance-based classifier based on the k-nearest neighbor (KNN) algorithm is considered a baseline technique for time series classification. 
Mostly, distance-based classifier uses Euclidean or Dynamic Time Warping (DTW) as a distance measure~\cite{keogh2005exact}. 
Feature-based time series classifiers are also widely used such as the bag-of-SFA-symbols (BOSS)~\cite{schafer2015boss} and the bag-of-features framework (TSBF)~\cite{baydogan2013bag} classifiers. 
Ensemble-based classifiers combine separate classifiers into one model to reach a higher classification accuracy such as the elastic ensemble (PROP)~\cite{lines2015time}, and the collective of transform-based ensemble (COTE)~\cite{bagnall2015time} classifiers. 

Convolutional neural network (CNN) based classifiers have advantages over other classification methods because CNNs provide the classifier with a preprocessing mechanism within the model. 
Examples are the multi-channel CNN (MC-CNN) classifier~\cite{cui2016multi}, the multi-layered preceptron (MLP)~\cite{wang2017time}, the fully convolutional network (FCN)~\cite{wang2017time} and, specifically,
the residual network (ResNet)~\cite{wang2017time}. 

The present paper focuses on
the recurrent neural network based classification approaches such as LSTM-FCN~\cite{karim2018lstm} and ALSTM-FCN~\cite{karim2018lstm}. 
These models combine both temporal CNNs and long short-term memory (LSTM) models to provide the classifier with both feature extraction and time dependencies through the dataset during the classification process. 
These models use additional support algorithms such as attention and fine-tuning algorithms to enhance the LSTM learning due to its complex structure and data requirements.

This paper studies whether the use of gated-recurrent units (GRUs) can improve
the hybrid classifiers listed above.
We create the GRU-FCN by replacing the LSTM with a GRU in the LSTM-FCN \cite{karim2018lstm}. We intentionally kept the other components of the entire model without changes to make an empirical comparison between the LSTM and GRU in a same model structure to obtain a fair comparison between both architectures regarding the univariate time series classification task.
%combined with a fully convolutional neural network (FCN) for time series classification which offers advantages over the LSTM-FCN~\cite{karim2018lstm}. We named the model GRU-FCN. 
Like the LSTM-FCN, our model does not require feature engineering or data preprocessing before the training or testing stages. 
The GRU is able to learn the temporal dependencies within the dataset. 
Moreover, the GRU has a smaller block architecture and shows comparable performance to the LSTM without need for additional algorithms to support the model. 

Although it is difficult to determine the best classifier for all time series types,
the proposed model seeks to achieve equivalent accuracy to state-of-the-art classification models in univariate time series classification. 
Following \cite{wang2017time} and \cite{karim2018lstm},
our tests use the UCR time series classification archive benchmark~\cite{ucr_benchmark} to compare our model with other state-of-the-art univariate time series classification models. Our model achieved higher classification performance on  several datasets
compared to other state-of-the-art classification models.

\begin{table}[t]%[!tb]
	\caption{Comparison of GRU and LSTM Computational Elements.} % title of Table
	\centering % used for centering table
	\begin{tabular}{ l  l  l  } % centered columns (4 columns)
		\textbf{Comparison} &\textbf{LSTM} & \textbf{GRU} \\
		\hline	\\
		number of gates	&3 & 2\\
		number of activations  &	2	&	1\\
		state memory cell &	Yes	&	No	\\
		number of weight matrices& 8 & 6 \\
		number of bias vectors&3  &  4  \\
		number of elementwise multiplies& 3 & 3 \\
		number of matrix multiplies& 8 & 6 \\
	\end{tabular}
	\label{table:gru_vs_lstm} % is used to refer this table in the text
\end{table}

\begin{figure}[t]
	\centering
	\includegraphics[width=8.5cm,height=6cm]{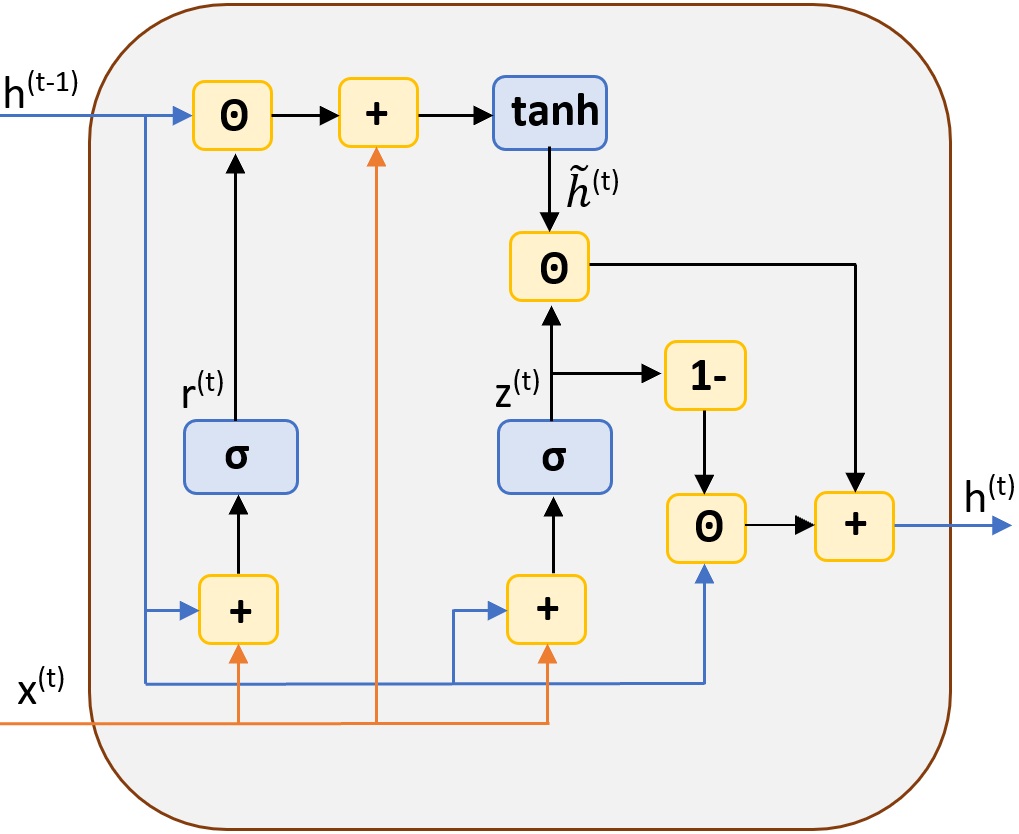}
	\caption{Block architecture for an unrolled GRU.}
	\label{gru_block}
\end{figure}

\begin{figure}
	\centering
	\includegraphics[width=8.5cm,height=20cm]{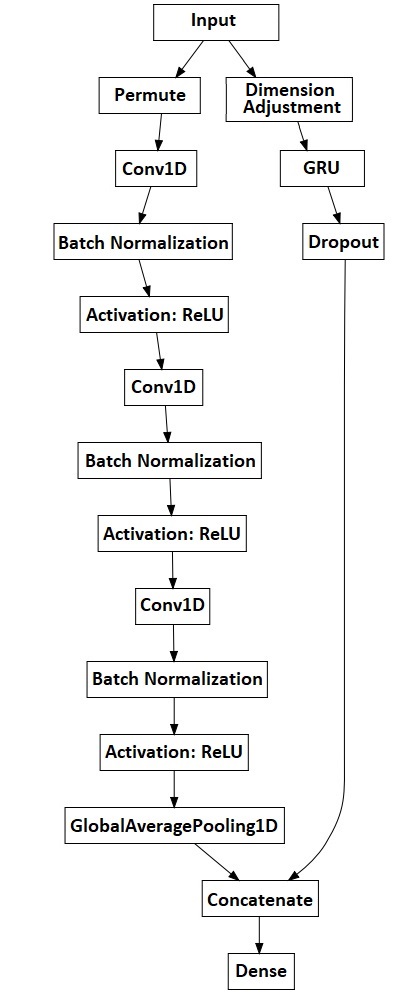}
	\caption{The proposed GRU-FCN model architecture diagram rendered using the Keras visualization tool and 
	modified from~\cite{karim2018lstm,wang2017time} architectures.}
	\label{model_architecture}
\end{figure}

%------Section Starts---------%
\section{Model Components}
%Our classification model combines gated recurrent unit (GRU) and fully convolutional neural network (FCN) which named as (GRU-FCN). 
%In this section, we give a brief introduction to the basic components of the proposed GRU-FCN model.
%------
\subsection{Gated Recurrent Unit (GRU)}
\begin{figure}
	\centering
	\includegraphics[width=8.5cm,height=6cm]{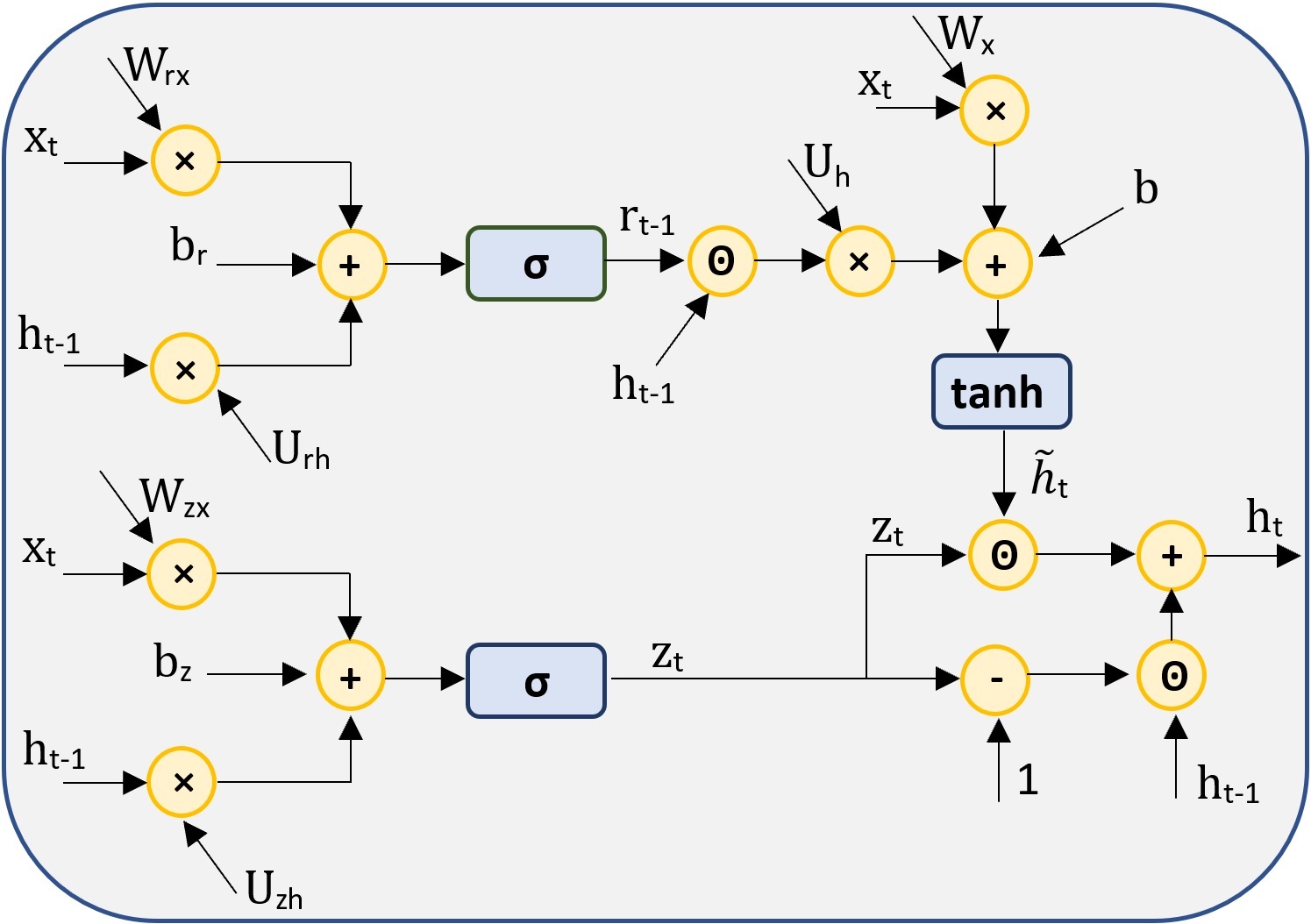}
	\caption{The GRU architecture showing the weights of each component.}
	\label{gru_weights}
\end{figure}

The gated recurrent unit (GRU) was introduced in~\cite{chung2014empirical} as another type of gate-based recurrent unit which has a smaller architecture and comparable performance to the LSTM unit. The GRU consists of two gates: reset and update. The architecture of an unrolled GRU block is shown in Figure.~\ref{gru_block}. $r^{(t)}$ and $z^{(t)}$ denote the values of the reset and update gates at time step $t$, respectively. 
$x_{i}\in\mathbb{R}^n$ is a $1\mathrm{D}$ input vector to the GRU block at time step $t$. ${\tilde h}^{(t)}$ is the output candidate of the GRU block. $h^{(t-1)}$ is the recurrent GRU block output of time step $t-1$ and the current output at time $t$ is $h^{(t)}$. 
Assuming a one-layer GRU, the reset gate, update gate, output candidate, and GRU output are calculated as follows~\cite{chung2014empirical}:
\begin{flalign}
%\begin{equation}
&z^{(t)} = \sigma (W_{zx} x^{(t)} + U_{zh} h^{(t-1)} + b_z) \label{eqn:z_gate}\\
%\end{equation}
%\begin{equation}
&r^{(t)} = \sigma (W_{rx} x^{(t)} + U_{rh} h^{(t-1)}+ b_r)\label{eqn:r_gate}\\
%\end{equation}
%\begin{equation}
&{\tilde h}^{(t)} = \mathrm{tanh} (W_{x} x^{(t)} + U_{h}( r^{(t)} \odot h^{(t-1)})+b)\label{eqn:h_hat}\\
%\end{equation}
%\begin{equation}
&h^{(t)} = (1-z^{(t)})\odot h^{(t-1)}+z^{(t)}\odot{\tilde h}^{(t)}\label{eqn:h}
%\end{equation}
\end{flalign}
Where $W_{zx}$, $W_{rx}$, and $W_{x}$ are the feedforward weights and $U_{hz}$, $U_{hr}$, and $U_{h}$ are the recurrent weights of the update gate, reset gate, and output candidate activation respectively. $b_z$, $b_r$ and $b$ are the biases of the update gate, reset gate and the output candidate activation ${\tilde h}^{(t)}$, respectively. 
Figure~\ref{gru_weights} shows the GRU architecture with weights and biases made explicit.

Like the RNN and LSTM, the GRU models temporal (sequential) datasets. 
The GRU uses its previous time step output and current input to calculate the next output. 
The GRU has the advantage of smaller size over the LSTM. 
The GRU consists of two gates (reset and update), while the LSTM has three gates: input, output and forget. 
The GRU has one unit activation, but the LSTM has two unit activations: input-update and output activations. 
Also, the GRU does not contain the memory state cell which exists in the LSTM model.
Thus, the GRU requires fewer trainable parameters, and  shorter training time compared to the LSTM. 
Table~\ref{table:gru_vs_lstm} compares GRU and LSTM architecture components. 

\subsection{Temporal Convolutional Neural Network}
The Convolutional Neural Network (CNN), introduced in 1989~\cite{lecun1989backpropagation}, 
utilizes weight sharing over grid-structured datasets such as images and time series~\cite{lecun1995convolutional, deepLearnigBook}. 
The convolutional layers within the CNN learn to extract complex feature representations from the data with little or no preprocessing. 
The temporal FCN consists of many layers of convolutional blocks that may have different or same kernel sizes, followed by a dense layer softmax classifier~\cite{deepLearnigBook}. 
For time series problems, the values of each convolutional block in the FCN, are calculated as follows~\cite{wang2017time}:
\begin{flalign}
%\begin{equation}
&y_{i} = W_{i}*x_{i}+b_{i}\\
%\end{equation}
%\begin{equation}
&z_{i} = \mathit{BN}(y)\\
%%\end{equation}
%\begin{equation}
&out_{i} = \mathit{ReLU}(z)
%\end{equation}
\end{flalign}
where $x_{i}\in\mathbb{R}^n$ is a $1\mathrm{D}$ input vector which represents a time series segment, $W_{i}$ is the $1\mathrm{D}$ convolutional kernel of weights, $b_{i}$ is the bias, and $y$ is the output vector of the convolutional block $i$. $z_{i}$ is the intermediate result 
after applying batch normalization~\cite{batch_normalization} on the convolutional block which then is passed to the rectified linear unit $ReLU$~\cite{relu} to calculate the output of the convolutional layer $out_{i}$.

\section{Model Architecture}
\begin{table*} 
	\begin{center}
		\renewcommand{\arraystretch}{0.7}
		\scriptsize
		\caption{The UCR datasets descriptions based on~\cite{ucr_benchmark} and their experimental adjustments used in the GRU-FCN implementation.}
		\begin{tabular}{p{2.2cm} p{1.5cm} p{1cm} p{1cm}p{1cm}p{1cm}p{1cm}p{1cm}p{0.7cm}}%{l l l l l l l l  l}
			
			\hline
			%	\textbf{Dataset}&\multicolumn{5}{c|}{\textbf{Description}} \\
			%\cline{2-6} 
			%	\multicolumn{}{| l |}
			\textbf{Dataset} &\textbf{\textit{Type}}& \textbf{\textit{\# Classes}}& \textbf{\textit{Length}}&\textbf{\textit{Train size}}&\textbf{\textit{Test size}} &\textbf{\textit{\# epochs}}&\textbf{\textit{Train Batch}}&\textbf{\textit{Test Batch}}\\
			%	\textbf{} & \textbf{}& \textbf{\textit{Number}}& \textbf{\textit{Size}}&\textbf{\textit{Size}}&\textbf{}\\
			\hline
			Adiac&	Image&	37&	176&	390&	391& 4000&128&128\\
			ArrowHead&	Image&	3&	251&	36&	175&4000&32&128\\
			Beef&	Spectro	&5	&470&	30&	30&8000&64&64\\
			BeetleFly&	Image&	2	&512&	20&	20&8000&64&64\\
			BirdChicken	&Image&	2	&512	&20	&20&8000&64&64\\
			Car	&Sensor&	4&	577&	60&	60&2000&128&128\\
			CBF	&Simulated&	3	&128	&30&	900&2000&32&128\\
			ChlorineConc&	Sensor&	3&	166	&467&	3840&2000&128&128\\
			CinCECGTorso&	Sensor&	4	&1639&	40	&1380&500&128&128\\
			Coffee	&Spectro	&2&	286&	28&	28&500&64&64\\
			Computers	&Device	&2	&720	&250	&250&2000&128&128\\
			CricketX	&Motion	&12&	300&	390	&390&2000&128&128\\
			CricketY	&Motion&	12&	300	&390&	390&2000&128&128\\
			CricketZ	&Motion&	12&	300	&390	&390&2000&64&128\\
			DiatomSizeR	&Image&	4	&345	&16	&306&2000&64&64\\
			DisPhOAgeGrp&	Image	&3	&80&	400	&139&2000&128&128\\
			DisPhOCorrect	&Image	&2	&80	&600	&276&2000&128&128\\
			DisPhTW	&Image&	6	&80	&400	&139&2000&128&128\\
			Earthquakes	&Sensor&	2	&512	&322	&139&2000&128&128\\
			ECG200	&ECG	&2	&96	&100	&100&8000&64&64\\
			ECG5000	&ECG	&5	&140&	500	&4500&2000&128&128\\
			ECGFiveDays	&ECG&	2	&136	&23	&861&2000&128&128\\
			ElectricDevices&	Device&	7	&96	&8926	&7711&2000&128&128\\
			FaceAll&	Image&	14	&131&	560&	1690&2000&128&128\\
			FaceFour	&Image	&4	&350	&24&	88&2000&128&128\\
			FacesUCR	&Image&	14&	131&	200	&2050&2000&128&128\\
			FiftyWords	&Image&	50&	270&	450	&455&2000&128&128\\
			Fish	&Image&	7	&463&	175	&175&2000&128&128\\
			FordA	&Sensor&	2&	500&	3601	&1320&2000&128&128\\
			FordB	&Sensor	&2	&500	&3636&	810&1600&128&128\\
			GunPoint	&Motion&	2	&150	&50	&150&2000&128&128\\
			Ham	&Spectro	&2	&431&	109	&105&2000&128&128\\
			HandOutlines	&Image&	2	&2709	&1000&	370&2000&64&128\\
			Haptics	&Motion&	5	&1092	&155&	308&2000&128&128\\
			Herring	&Image&	2	&512&	64	&64&2000&128&128\\
			InlineSkate&	Motion&	7	&1882&	100&	550&2000&128&128\\
			InsWingSound	&Sensor	&11	&256	&220	&1980&1000&128&128\\
			ItalyPowD	&Sensor	&2&	24&	67	&1029&2000&64&128\\
			LargeKApp	&Device&	3	&720	&375	&375&2000&128&128\\
			Lightning2	&Sensor&	2	&637	&60	&61&4000&128&128\\
			Lightning7	&Sensor&	7&	319&	70	&73&3000&32&32\\
			Mallat&	Simulated	&8&	1024&	55	&2345&2500&128&128\\
			Meat	&Spectro	&3	&448&	60	&60&2000&64&128\\
			MedicalImages	&Image	&10	&99	&381	&760&2000&64&128\\
			MidPhOAgeGrp&	Image&	3&	80&	400	&154&2000&128&128\\
			MidPhOCorrect&	Image&	2	&80&	600	&291&2000&128&128\\
			MidPhTW&Image	&6	&80&	399	&154&2000&128&128\\
			MoteStrain	&Sensor	&2&	84	&20&	1252&2000&128&128\\
			NonInvECGTh1	&ECG	&42	&750	&1800&	1965&2000&128&128\\
			NonInvECGTh2	&ECG	&42	&750	&1800&	1965&2000&128&128\\
			OliveOil&	Spectro	&4	&570&	30	&30&6000&64&128\\
			OSULeaf	&Image	&6	&427&	200	&242&2000&64&128\\
			PhalOCorrect	&Image&	2&	80&	1800	&858&2000&64&128\\
			Phoneme&	Sensor	&39	&1024	&214&	1896&2000&64&128\\
			Plane	&Sensor	&7	&144	&105	&105&200&16&16\\
			ProxPhOAgeGrp	&Image&	3	&80&	400	&205&2000&128&128\\
			ProxPhOCorrect	&Image	&2&	80	&600&	291&2000&128&128\\
			ProxPhTW	&Image&	6	&80&	400	&205&2000&128&128\\
			RefDevices	&Device	&3	&720	&375	&375&2000&64&64\\
			ScreenType	&Device&	3	&720	&375	&375&2000&64&128\\
			ShapeletSim	&Simulated	&2	&500&	20	&180&2000&128&128\\
			ShapesAll	&Image	&60	&512&	600	&600&4000&64&64\\
			SmlKitApp&	Device	&3&	720	&375&	375&2000&128&64\\
			SonyAIBORI&	Sensor	&2	&70&	20&	601&2000&64&128\\
			SonyAIBORII	&Sensor&	2	&65	&27&	953&2000&64&128\\
			StarLightCurves&	Sensor	&3	&1024	&1000	&8236&2000&64&64\\
			Strawberry&	Spectro	&2	&235&	613	&370&8000&64&64\\
			SwedishLeaf	&Image	&15	&128&	500	&625&8000&64&64\\
			Symbols	&Image&	6	&398&	25&	995&2000&64&64\\
			SynControl&	Simulated&	6&	60&	300	&300&4000&16&128\\
			ToeSegI&	Motion&	2	&277&	40	&228&2000&128&64\\
			ToeSegII	&Motion&	2	&343&	36	&130&2000&128&32\\
			Trace&	Sensor	&4	&275	&100&	100&1000&64&128\\
			TwoLeadECG	&ECG	&2&	82&	23	&1139&2000&64&64\\
			TwoPatterns	&Simulated&	4	&128	&1000&	4000&2000&32&128\\
			UWaveAll&	Motion&	8	&945&	896	&3582&500&16&16\\
			UWaveX&	Motion	&8&	315&	896&	3582&2000 &64& 16\\
			UWaveY&	Motion	&8	&315&	896	&3582&2000& 64& 64\\
			UWaveZ	&Motion	&8	&315&	896&	3582& 2000 &64& 64 \\
			Wafer	&Sensor	&2	&152	&1000&	6164 &1500 &64 &64\\
			Wine	&Spectro	&2	&234&	57	&54&8000&64&64\\
			WordSynonyms	&Image	&25&	270	&267	&638 & 1500& 64 &64\\
			Worms	&Motion	&5	&900	&181&	77&2000&64&64\\
			WormsTwoClass	&Motion	&2&	900	&181&	77&1000&16&16\\
			Yoga	&Image	&2	&426&	300&	3000 &1000 &128& 128\\
			\hline
			\label{dataset_desctiption}
		\end{tabular}
	\end{center}
\end{table*}

As stated in the introduction, our model replaces the LSTM with a GRU in a hybrid gated-FCN. We intentionally did not change the other components of the entire model to attain a fair comparison between GRU and LSTM architectures in the same model structure for univariate time series classification. 
Our model is based on the framework introduced in~\cite{wang2017time, karim2018lstm}. 
The proposed architecture actual implementation is shown in Figure~\ref{model_architecture}.
The architecture has two parallel parts: a GRU and a temporal FCN. 
Our model uses three layers FCN architecture proposed in~\cite{wang2017time}. The dimension adjustment aims to change the dimensions of the input to be compatible with the GRU recurrent design~\cite{keras_rnns}. We also used the global average pooling layer~\cite{boureau2010theoretical} to interpret the classes and to reduce the number of trainable parameters comparing to the fully connected layer, without any sacrifice in the accuracy. 
The FCN $\mathrm{1D}$ kernel numbers are $128$, $256$, and $128$ with kernel sizes $8$, $5$, and $3$ in each convolutional layer, respectively. The weights were initialized using the He uniform variance scaling initializer~\cite{he2015delving}. In addition, we used the GRU instead of LSTMs that were used in~\cite{karim2018lstm} models to reduce the number of trainable parameters, memory, and training time. 
Moreover, we removed the masking and any extra supporting algorithms such as an attention mechanism, and fine-tuning that were used in the LSTM-FCN and ALSTM-FCN models~\cite{karim2018lstm}. The GRU is unfolded by eight unfolds as used in~\cite{karim2018lstm} for univariate time series. 
The hyperbolic tangent ($\mathit{tanh}$) function used as the unit activation and 
the hard-sigmoid ($\mathit{hardSig}$) function~\cite{Gulcehre2016} is used as the recurrent activation (gate activation) of the GRU architecture. The weights were initialized using the $\mathit{glorot\_uniform}$ initializer~\cite{glorot2010understanding, kerasAPI} and the biases were initialized to zero. 
The input was fitted using the concept used in~\cite{karim2018lstm} to fit an input to a recurrent unit. We used the Adam optimization function~\cite{kingma2014adam} with $\beta_1 = 0.9$, $\beta_2 = 0.999$ and initial learning rate $\alpha= 0.01$. The learning rate $\alpha$ was reduced by a factor of $0.8$ every $100$ training steps until it reached the minimum rate $\alpha= 0.0001$. 
The dense layer uses the softmax classifier~\cite{nasrabadi2007pattern} using the categorical crossentropy loss 
function~\cite{deepLearnigBook}. 
In this paper our goal is to make a fair comparison between the LSTM-based model and our GRU-based model.
Thus, we used the same number of epochs that was assigned by the original LSTM-FCN model~\cite{karim2018lstm} for each univariate time series. The number of epochs that we assigned for each dataset used is shown in Table~\ref{dataset_desctiption} .

The input to the model is the raw dataset without applying any normalizations or feature engineering prior to the training process. 
The FCN is responsible for feature extraction from the time series~\cite{wang2017time} and the GRU enables the model to learn temporal dependencies within the time series. 
Therefore the model learns both the features and temporal dependencies to predict the correct class for each training example. 

\begin{figure*}[h]
	\centering
	\includegraphics[width=14cm,height=6cm]{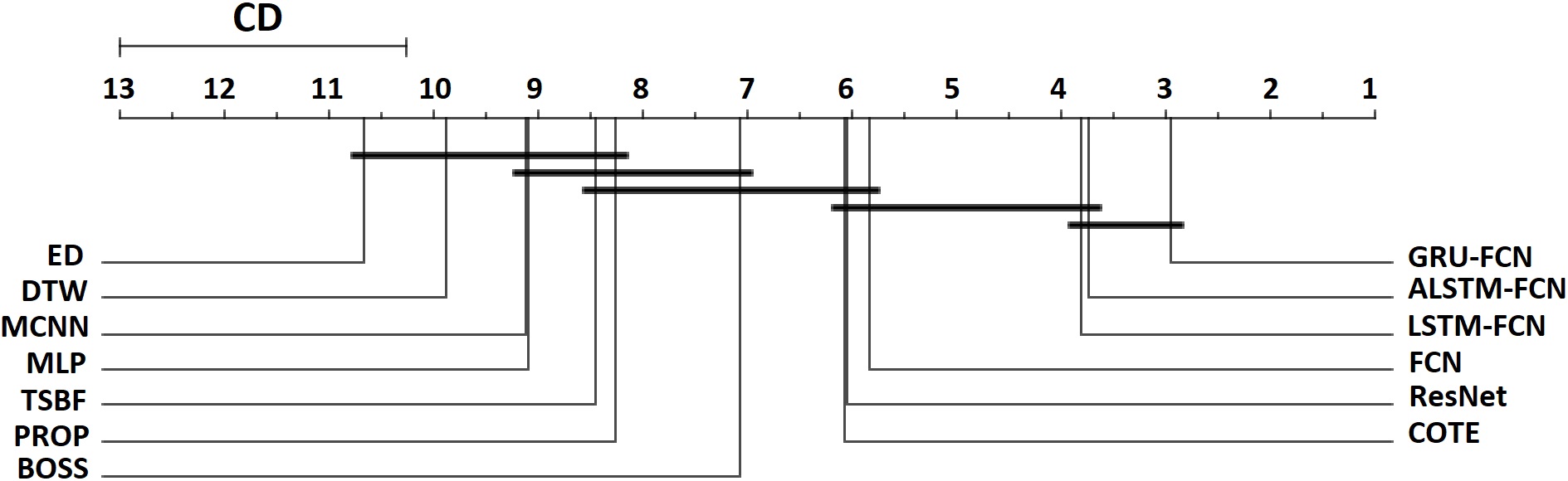}
	\caption{Critical difference diagram based on arithmetic mean of model ranks.}
	\label{critical_diagram}
\end{figure*}

\section{Method and Results}

\begin{table*}
	\renewcommand{\arraystretch}{0.8}
	\caption{Classification testing error and rank for 85 time series datasets from the UCR benchmark.}
	\scriptsize%\footnotesize
	\begin{center}
		\begin{tabular}{|l|lllllllllllll|}%{p{1.2cm} p{1cm}p{1cm}p{1cm}p{1cm}p{1cm}p{1cm}p{1cm}p{1cm}p{1cm}p{1cm}p{1cm}p{1cm}p{1cm}p{1cm}}
			\hline
			\textbf{Dataset}&\multicolumn{13}{|c|}{\textbf{Classification Method and Testing Error}} \\
			\cline{2-14} 
			\textbf{} & \textbf{\textit{GRU-FCN}}& \textbf{\textit{FCN}}& \textbf{\textit{LSTMFCN}}&\textbf{\textit{ALSTMFCN}}&\textbf{\textit{ResNet}}&\textbf{\textit{MCNN}}&\textbf{\textit{MLP}}&\textbf{\textit{COTE}}&\textbf{\textit{DTW}}&\textbf{\textit{PROP}}&\textbf{\textit{BOSS}}&\textbf{TSBF}&\textbf{\textit{ED}}\\
			\hline
			Adiac&\textbf{0.127}&0.143&0.141&0.139&0.174&0.231&0.248&0.233&0.396&0.353&0.235&0.231&0.389\\
			ArrowHead&\textbf{0.085}&0.120&0.102&0.119&0.183&/&0.292&0.138&0.297&0.103&1.66&0.246&0.200\\
			Beef&\textbf{0.100}&0.250&0.167&0.233&0.233&0.367&0.167&0.133&0.367&0.367&0.200&0.434&0.333\\
			BeetleFly&\textbf{0.050}&\textbf{0.050}&\textbf{0.050}&\textbf{0.050}&0.200&/&0.200&0.050&0.300&0.400&0.100&0.200&0.250\\
			BirdChicken&\textbf{0}&0.050&\textbf{0}&\textbf{0}&0.100&/&0.400&0.150&0.250&0.350&0.050&0.100&0.450\\
			Car&\textbf{0.016}&0.050&0.033&0.159&0.067&/&0.117&/&0.267&/&0.167&0.217&0.267\\
			CBF&\textbf{0}&0.008&0.003&0.004&0.006&0.002&0.14&0.001&0.003&0.002&0.002&0.013&0.148\\
			ChloConc&\textbf{0.002}&0.157&0.191&0.193&0.172&0.203&0.125&0.314&0.352&0.360&0.339&0.308&0.350\\
			CinCECGTorso&0.124&0.187&0.191&0.193&0.172&0.058&0.158&\textbf{0.064}&0.349&0.062&0.125&0.288&0.103\\
			Coffee&\textbf{0}&\textbf{0}&\textbf{0}&\textbf{0}&\textbf{0}&0.036&\textbf{0}&\textbf{0}&\textbf{0}&\textbf{0}&\textbf{0}&\textbf{0}&\textbf{0}\\
			Computers&0.148&0.152&0.136&0.123&0.176&/&0.504&0.240&0.300&\textbf{0.116}&0.244&0.244&	0.424\\
			CricketX&0.156&0.185&0.193&0.203&0.179&0.182&0.431&\textbf{0.154}&0.246&0.203&0.259&0.295&0.423\\
			CricketY&0.156&0.208&0.183&0.185&0.195&\textbf{0.154}&0.405&0.167&0.256&0.156&0.208&0.265&0.433\\
			Cricketz&0.154&0.187&0.190&0.175&0.169&0.142&0.408&\textbf{0.128}&0.246&0.156&0.246&0.285&0.413\\
			DiatomSizeR&0.036&0.069&0.046&0.063&0.069&\textbf{0.023}&0.036&0.082&0.033&0.059&0.046&0.102&0.065\\
			DisPhOAgeGr&0.142&0.165&0.145&\textbf{0.137}&0.202&/&0.178&0.229&0.230&0.223&0.272&0.218&0.374\\
			DisPhOCorrect&0.168&0.188&0.168&\textbf{0.163}&0.180&/&0.195&0.238&0.283&0.232&0.252&0.288&0.283\\
			DisPhalanxTW&\textbf{0.180}&0.210&0.185&0.185&0.260&/&0.375&0.317&0.410&0.317&0.324&0.324&0.367\\
			Earthquakes&\textbf{0.171}&0.199&0.177&0.173&0.214&/&10.208&/&0.281&0.281&0.186&0.252&0.288\\
			ECG200&\textbf{0.080}&0.100&0.100&0.090&0.130&/&0.210&0.150&0.230&/&0.130&0.160&0.120\\
			ECG5000&\textbf{0.052}&0.059&0.053&\textbf{0.052}&0.069&/&0.068&0.054&0.076&0.350&0.059&0.061&0.075\\
			ECG5Days&\textbf{0}&0.010&0.011&0.009&0.045&\textbf{0}&0.030&\textbf{0}&0.232&0.178&\textbf{0}&0.124&0.203\\
			ElectricDevices&\textbf{0.037}&0.277&\textbf{0.037}&\textbf{0.037}&0.272&/&0.360&0.230&0.399&0.277&0.201&0.298&0.449\\
			FaceAll&\textbf{0.040}&0.071&0.060&0.045&0.166&0.235&0.115&0.105&0.192&0.115&0.210&0.256&0.286\\
			FaceFour&0.136&0.068&0.057&0.057&0.068&0&0.167&0.091&0.171&0.091&\textbf{0}&\textbf{0}&0.216\\
			FourUCR&0.050&0.052&0.071&0.057&\textbf{0.042}&0.063&0.185&0.057&0.095&0.063&\textbf{0.042}&0.134&0.231\\
			FiftyWords&\textbf{0.167}&0.321&0.196&0.176&0.273&0.190&0.288&0.191&0.301&0.180&0.301&0.242&0.369\\
			Fish&\textbf{0.006}&0.029&0.017&0.023&0.011&0.051&0.126&0.029&0.177&0.034&0.011&0.166&0.217\\
			FordA&0.074&0.094&\textbf{0.072}&0.073&\textbf{0.072}&/&0.231&/&0.444&0.182&0.083&0.150&0.335\\
			FordB&0.083&0.117&0.088&\textbf{0.081}&0.100&/&0.371&/&0.380&0.265&0.109&0.402&0.394\\
			GunPoint&\textbf{0}&\textbf{0}&\textbf{0}&\textbf{0}&0.007&\textbf{0}&0.067&0.007&0.093&0.007&\textbf{0}&0.014&0.087\\
			Ham&0.209&0.238&0.209&0.228&0.219&/&\textbf{0.162}&0.334&0.533&/&0.334&0.239&0.400\\
			HandOutlines&0.112&0.224&0.113&0.358&0.139&/&0.117&0.068&0.119&/&\textbf{0.098}&0.146&0.138\\
			Haptics&0.455&0.449&\textbf{0.425}&0.435&0.495&0.530&0.539&0.488&0.623&0.584&0.536&0.510&0.630\\
			Herring&0.250&0.297&0.250&0.265&0.406&/&0.360&0.313&0.469&\textbf{0.079}&0.454&0.360&0.484\\
			InlineSkate&0.625&0.589&0.534&\textbf{0.507}&0.635&0.618&0.649&0.551&0.616&0.567&0.511&0.615&0.658\\
			InsWSound&0.446&0.598&0.342&\textbf{0.329}&0.469&/&0.369&/&0.643&/&0.479&0.376&0.438\\
			ItalyPower&\textbf{0.027}&0.030&0.037&0.040&0.040&0.030&0.034&0.036&0.050&0.039&0.053&0.117&0.045\\
			LKitApp&0.090&0.104&0.090&\textbf{0.083}&0.107&/&0.520&0.136&0.205&0.232&0.235&0.472&0.507\\
			Lightening2&0.197&0.197&0.197&0.213&0.246&0.164&0.279&0.164&0.131&\textbf{0.115}&0.148&0.263&0.246\\
			Lightening7&\textbf{0.137}&\textbf{0.137}&0.164&0.178&0.164&0.219&0.356&0.247&0.274&0.233&0.342&0.274&0.427\\
			MALLAT&0.048&0.020&0.019&\textbf{0.016}&0.021&0.057&0.064&0.036&0.066&0.050&0.058&0.040&0.086\\
			Meat&0.066&0.033&0.116&0.033&\textbf{0}&/&\textbf{0}&0.067&0.067&/&0.100&0.067&0.067\\
			MedicalImages&\textbf{0.199}&0.208&\textbf{0.199}&0.204&0.228&0.260&0.271&0.258&0.263&0.245&0.288&0.295&0.316\\
			MidPhOAgeGrp&0.187&0.232&0.188&0.189&0.240&/&	0.193&\textbf{0.169}&	0.500&	0.474&	0.220&	0.186&	0.481\\
			MidPhOCorrect&	\textbf{0.160}&	0.205&	\textbf{0.160}&	0.163&	0.207&	/&	0.442&	0.403&	0.302&	0.210&	0.455&	0.423&	0.234\\
			MidPhTW&	\textbf{0.363}&	0.388&	0.383&	0.373&	0.393&	/&	0.429&	0.429&	0.494&	0.630&	0.455&	0.403&	0.487\\
			MoteStrain&	0.076&\textbf{0.050}&	0.061&	0.064&	0.105&	0.079&	0.131&	0.085&	0.165&	0.114&	0.073&	0.097&	0.121\\
			NonInvECGTh1 &	\textbf{0.034}&	0.039&	0.035&	0.025&	0.052&	0.064&	0.058&	0.093&	0.210&	0.178&	0.161&	0.158&	0.171\\
			NonInvECGTh2 &	\textbf{0.035}&	0.045	&0.038&	0.034&	0.049&	0.060&	0.057&	0.073&	0.135&	0.112&	0.101&	0.139&	0.120\\
			OliveOil&	\textbf{0.012}&	0.167&	0.133&	0.067&	0.133&	0.133&	0.600&	0.100&	0.167&	0.133&	0.100&	0.167&	0.133\\
			OSULeaf&\textbf{0}&	0.012&	0.004&	0.004&	0.021&	0.271&	0.430&	0.145&	0.409&	0.194&	0.012&	0.240&	0.479\\
			PhalOCorrect&	\textbf{0.165}&	0.174&	0.177&	0.170&	0.175&	/&	0.164&	0.194&	0.272&	/&	0.229&	0.171&	0.239\\
			Phoneme&	0.644&	0.655&	0.650&	\textbf{0.640}&	0.676&	/&	0.902&	/&	0.772&	/&	0.733&	0.724&	0.891\\
			Plane&	\textbf{0}&	\textbf{0}&	\textbf{0}&	\textbf{0}&	\textbf{0}&	/&	0.019&/	&	\textbf{0}&/&	/&	\textbf{0}&	0.038\\
			ProxPhOeAgeGrp&	0.117&	0.151&	0.117&	\textbf{0.107}&	0.151&/&	0.135&	0.121&	0.195&	0.117&	0.152&	0.128&	0.215\\
			ProxPhOCorrect&	0.079&	0.100&	\textbf{0.065}&	0.075&	0.082&/&	0.200&	0.142&	0.217&	0.172&	0.166&	0.152&	0.192\\
			ProxPhTW&	\textbf{0.167}&	0.190&	\textbf{0.167}&	0.173&	0.193&/&	0.210&	0.186&	0.244&	0.244&	0.200&	0.191&	0.293\\
			RefDevices	&\textbf{0.407}&	0.467&	0.421&	0.429&	0.472&/&	0.632&	0.443&	0.536&	0.424&	0.498&	0.528&	0.605\\
			ScreenType&	0.297&	0.333&	0.351&	0.341&\textbf{0.293}&/&	0.614&	0.411&	0.603&	0.440&	0.536&	0.491&	0.640\\
			ShapeletSim&	0.011&	0.133&	0.011&	0.011&\textbf{0}&/&	0.528&\textbf{0}&	0.350&	/&	\textbf{0}&	0.039&	0.461\\
			ShapesAll&	0.097&	0.102&	0.098&	0.100&	\textbf{0.088}&/&	0.350&	0.095&	0.232&	0.187&	0.092&	0.815&	0.248\\
			SmlKitApp&	0.186&	0.197&	0.184&	0.203&	0.203&	/&	0.667&	\textbf{0.147}&	0.357&	0.187&	0.275&	0.328&	0.659\\
			SonyAIBORI	&0.017&	0.032&	0.018&	0.030&	\textbf{0.015}&	0.230&	0.273&	0.146&	0.275&	0.293&	0.321&	0.205&	0.305\\
			SonyAIBORII &	\textbf{0.018}&	0.038&	0.022&	0.025&	0.038&	0.070&	0.161&	0.076&	0.169&	0.124&	0.098&	0.223&	0.141\\
			StarLightCurves &	0.025&	0.033&	0.024&	0.023&	0.029&	0.023&	0.043&	0.031	&0.093&	0.079&	\textbf{0.021}&	0.023&	0.151\\
			Strawberry&	\textbf{0.013}&	0.031&\textbf{0.013}&\textbf{0.013}&0.042&	/&	0.038&	0.030&	0.059&	/&	0.025&	0.046&	0.054\\
			SwedishLeaf &	0.016&	0.034&	0.021&\textbf{0.014}&	0.042&	0.066&	0.107&	0.046&	0.208&	0.085&	0.272&	0.085&	0.211\\
			Symbols&	0.024&	0.038&	0.016&	\textbf{0.013}&	0.128&	0.049&	0.147&	0.046&	0.050&	0.049&	0.032&	0.055&	0.101\\
			SynControl& \textbf{0}&	0.010&	0.003&	0.006&	\textbf{0}&	0.003&	0.050&	\textbf{0}&	0.007&	0.010&	0.030&	0.007&	0.120\\
			ToeSeg1&	0.021&	0.031&	\textbf{0.013}&	\textbf{0.013}&	0.035&/&	0.500&	0.018&	0.228&	0.079&	0.062&	0.220&	0.320\\
			ToeSeg2	&0.076	&0.085&	0.084	&0.077&	0.138&	/&	0.408&	0.047&	0.162&	0.085&\textbf{0.039}&	0.200&	0.192\\
			Trace &	\textbf{0}&	\textbf{0}&	\textbf{0}&	\textbf{0}&	\textbf{0}&	\textbf{0}&	0.180&	0.010&	\textbf{0}&	0.010&	\textbf{0}&	0.020&	0.240\\
			TwoLeadECG&	\textbf{0}&	\textbf{0}&	0.001&	0.001&	\textbf{0}&	0.001&	0.147&	0.015&	0.096&	\textbf{0}&	0.004&	0.135&	0.253\\
			TwoPatterns	&0.009&	0.103&	0.003&	0.003&	\textbf{0}&	0.002&	0.114&	\textbf{0}&	\textbf{0}&	0.067&	0.016&	0.024&	0.093\\
			UWaveAll&	0.078&	0.174&	0.096	&0.107	&0.132&/&	0.253&	0.161&	0.108&	0.199&	0.238&	0.170&	\textbf{0.052}\\
			UWaveX &	0.171&	0.246&	\textbf{0.151}&	0.152&	0.213&	0.180&	0.232&	0.196&	0.273&	0.199&	0.241&	0.264&	0.261\\
			UWaveY&	0.240&	0.275&	\textbf{0.233}&	0.234&	0.332&	0.268&	0.297	&0.267&	0.366&	0.283&	0.313	&0.228&	0.338\\
			UWaveZ&	0.237&	0.271&	0.203&	\textbf{0.202}&	0.245&	0.232&	0.295&	0.265&	0.342&	0.290&	0.312&	0.074&	0.350\\
			Wafer &	\textbf{0.001}&	0.003&	\textbf{0.001}&	0.002&	0.003&	0.002&	0.004&\textbf{0.001}&	0.020&	0.003&	\textbf{0.001}&	0.005&	0.005\\
			Wine&	\textbf{0.111}&	\textbf{0.111}&	\textbf{0.111}&	\textbf{0.111}&	0.204&/&0.056&	0.223&	0.426&/&	0.260&	0.389&	0.389\\
			WordSynonyms &	0.262&	0.420&	0.329&	0.332&	0.368&	0.276&	0.406&	0.266&	0.351&	\textbf{0.226}&	0.345&	0.312&	0.382\\
			Worms&	0.325&	0.331&	0.325&	\textbf{0.320}&	0.381&	/&	0.585&	0.442&	0.416&	/&	0.442&	0.312&	0.545\\
			WormsTwoClass	&0.209&	0.271&	0.226&	0.198&	0.265&/&	0.403&	0.221&	0.377&/&	\textbf{0.169}&	0.247&	0.390\\
			Yoga &	0.090&	0.098&	0.082&	\textbf{0.081}&	0.142&	0.112&	0.145&	0.113&	0.164&	0.121&	\textbf{0.081}&	0.181&	0.170\\
			
			\Xhline{2\arrayrulewidth}
			\textbf{no. best}&\textbf{39}&9&19&25&13&5&3&11&4&5&13&3&2\\
			\textbf{Arith AVG Rank}&\textbf{2.947}	&5.841&	3.818&	3.729&	6.035&	9.118&	9.100&	6.071&	9.882&	8.253&	7.071&	8.459&	10.676\\
			\textbf{MPCE}&\textbf{0.0308}&	0.0387&	0.0327&	0.0342	&0.0415&0.1853	&0.0725&	0.0629&	0.0734&	0.1018&	0.0558&	0.0599&	0.0807\\
			\hline
		\end{tabular}
		\label{table1}
	\end{center}
\end{table*}

We implemented our model by modifying the original LSTM-FCN~\cite{karim2018lstm}. We found that the fine-tuning algorithm has not been applied in the actual LSTM-FCN and ALSTM-FCN implementation on source code github which shared by the authors~\cite{karim2018lstm} and mentioned in their literature. In addition, the LSTM-FCN~\cite{karim2018lstm} authors used a permutation algorithm for fitting the input to the FCN part which was not mentioned in their literature. Therefore, we generated the actual LSTM-FCN and ALSTM-FCN implementations to record the results based on their actual code implementation. In addition, to record their training time, memory requirement, number of parameters and f1-score. The Keras API~\cite{kerasAPI} with TensorFlow backend~\cite{tensorflow2015-whitepaper} were used in the implementation of the LSTM-FCN, ALSTM-FCN and GRU-FCN models. The source code of our GRU-FCN implementation can be found on github:~{https://github.com/NellyElsayed/GRU-FCN-model-for-univariate-time-series-classification}.

We tested our model on the UCR time series archive~\cite{ucr_benchmark} as one of the standard benchmarks for time series classification. Each dataset is divided into training and testing sets. The number of classes in each time series, the length and the size of both the training and test sets are shown in Table~\ref{dataset_desctiption} based on the datasets description in~\cite{ucr_benchmark}.
The UCR benchmark datasets have different types of collected sources: $29$ datasets of image source, $6$ spectro source, $5$ simulated source, $19$ sensor source, $6$ device source, $12$ motion source, and $6$ electrocardiogram (ECG) source. In addition, as we mentioned in the previous Section, Table~\ref{dataset_desctiption} also shows the number of epochs through training, and the batch sizes of the training and testing stages based on our experiments.

We compared our GRU-FCN with several state-of-the-art time series methods that also were studied in~\cite{wang2017time} and~\cite{karim2018lstm}. These included  FCN~\cite{wang2017time} which is based on 
a fully convolutional network, LSTM-FCN~\cite{karim2018lstm}, ALSTM-FCN~\cite{karim2018lstm}, 
that are based on 
long short-term memory and fully convolutional networks, ResNet \cite{wang2017time} which based on convolutional residual networks, multi-scale convolution neural networks model (MCNN)~\cite{cui2016multi}, multi-layered perceptrons model (MLP)~\cite{wang2017time}, collective of transformation-based ensembles model (COTE)~\cite{bagnall2015time} which based on transformation ensembles, dynamic time warping model (DTW)~\cite{jeong2011weighted} that is based on a weighted dynamic time warping mechanism, PROP model~\cite{lines2015time} which is based on elastic distance measures, BOSS model~\cite{schafer2015boss} that based on noise reduction in the time series representation,  time series based on a bag-of-features representation (TSBF) model~\cite{baydogan2013bag}, and Euclidean distance (ED) model~\cite{ucr_benchmark}. 
Our model shows the overall highest number of being the best classifier for $\mathrm{39}$ time series out of $\mathrm{85}$. 
Our model also shows the overall smallest classification error, arithmetic average rank, and mean per-class 
classification error (MPCE) compared to the other models as shown in Table~\ref{table1}.

Table~\ref{parameters_compare} shows a comparison between the number of parameters, training time and memory required to save the trainable weights of the GRU-FCN and both LSTM-FCN and ALSTM-FCN models as the existing LSTM-based to-date univariate classification models over the UCR $\mathrm{85}$ datasets. The GRU-FCN has smaller number of parameters for all the datasets. The GRU-FCN saves overall $\mathrm{1207 KB}$, and $\mathrm{5719 KB}$ memory requirements to save the trained model's weight; and $\mathrm{106.065}$, and $\mathrm{62.271}$  minutes to train the models over the UCR datasets comparing to the LSTM-FCN and ALSTM-FCN, respectively. Therefore, the GRU-FCN is preferable as low budget classification model with high accuracy performance.

We evaluated our model using the Mean Per-Class Error (MPCE) used in~\cite{wang2017time} to evaluate performance of a classification method over multiple datasets. 
The MPCE for a given model is calculated based on the per-class error (PCE) as follows:
\begin{flalign}
&\mathrm{PCE}_m = \frac{e_m}{c_m}\\
&\mathrm{MPCE} = \frac{1}{M} \sum_{m=1}^{M}\mathrm{PCE}_m
\end{flalign}
where $e_m$ is the error rate for dataset $m$ consisting of $c_m$ classes.
$M$ is the number of tested datasets.

\begin{table*} 
	\begin{center}
		\renewcommand{\arraystretch}{0.9}
		\scriptsize
		\caption{A comparison between the GRU-FCN and LSTM-based classification models for number of parameters, training time (minutes), and memory (KB) required to save the model weights on the UCR $\mathrm{85}$ datasets~\cite{ucr_benchmark}.}
		\begin{tabular}{l| l l l |rrr |rrr}%{p{2.2cm} p{1.5cm}p{1cm}p{1cm}p{1cm}p{1cm}p{1cm}p{1cm}p{0.7cm}p{0.7cm}}
			
			\hline
			\textbf{Dataset}&\multicolumn{3}{|c|}{\textbf{Number of Parameters}} &\multicolumn{3}{|c|}{\textbf{Training Time (Minutes)}}&\multicolumn{3}{|c}{\textbf{Memory (KB)}}\\
			\cline{2-10} 
			%	\multicolumn{}{| l |}
		%	\hline
			\textbf{} &\textbf{\textit{GRU-FCN}}& \textbf{\textit{LSTM-FCN}}& \textbf{\textit{ALSTM-FCN}}&\textbf{\textit{GRU-FCN}}&\textbf{\textit{LSTM-FCN}} &\textbf{\textit{ALSTM}}&\textbf{\textit{GRU-FCN}}&\textbf{\textit{LSTM-FCN}}&\textbf{\textit{LSTM-FCN}}\\
			%	\textbf{} & \textbf{}& \textbf{\textit{Number}}& \textbf{\textit{Size}}&\textbf{\textit{Size}}&\textbf{}\\
			\hline
			Adiac&275,237&276,717&283,837&9.597&9.560&10.056&1,114&1,119&1,150\\
			ArrowHead&272,379&274,459&284,579&4.134&4.303&4.692&1,103&1,111	&1,151\\
			Beef&277,909&281,741&300,621&3.896&4.804&4.889&1,124&1,139&1,215\\
			BeetleFly&278,506&282,674&303,234&3.937&4.208&4.545&1,126&1,144	&1,225\\
			BirdChicken&278,506&282,674	&303,234&3.437&3.760&4.131&1,126&1,144&1,225\\
			Car&280,340&285,028&308,188&1.899&1.972&2.045&1,134&1,152&1.245\\
			CBF&269,427&270,523&275,723&5.243&5.248&5.339&1,092&1,096&1,117\\
			ChloConc&270,339&271,739&278,459&13.324&14.601&14.813&1,095&1,110&1,127\\
			CinCECGTorso&305,828&319,012&384,652&6.087&6.594&7.003&1,233&1,285&1,544\\
			Coffee&273,082&275,442&286,962&0.504&0.524&0.540&1,104&1,115&1,161\\
			Computers&283,498&289,330&318,210&7.722&8.049&8.436&1,145&1,170	&1,283\\
			CricketX&274,788&277,260&289,340&6.850&7.124&7.292&1,112&1,122&1,171\\
			CricketY&274,788&277,260&289,340&6.673&6.978&7.224&1,112&1,122&1,171\\
			Cricketz&274,788&277,260&289,340&8.601&8.933&9.539&1,112&1,122&1,171\\
			DiatomSizeR&274,772&277,604&291,484&2.886&3.016&3.066&1,112&1,123&1,180\\
			DisPhOAgeGrp&268,275&268,987&272,267&2.346&2.439&5.056&1,087&1,090&1,103\\
		    DisPhOCorrect&268,138&268,850&272,130&3.554&3.791&3.980&1,085&1,090	&1,103\\
		    DisPhTW&268,686&269,398&272,678&2.611&2.723&2.876&1,088&1,091&1,106\\
	    	Earthquakes&278,506&282,674&303,234&4.998&5.507&5.547&1,126&1,144&1,225\\
	    	ECG200&268,522&269,362&273,282&5.305&5.599&6.125&1,087&1,092&1,108\\
		    ECG5000&269,989&271,181&276,861&13.223&13.797&14.162&1,093&1,098&1,123\\
		    ECG5Days&269,482&270,642&276,162&2.433&2.481&2.494&1,090&1,097&1,119\\
		    ElectricDevices&269,207&270,047&273,967&67.350&75.44&65.879&1,090&1,093&1,111\\
		    FaceAll&271,006&272,126&277,446&7.465&7.753&7.812&1,097&1,101&1,125\\
		    FaceFour&274,892&277,764&291,844&1.072&1.101&1.197&1,112&1,123&1,181\\
		    FourUCR&271,006&272,126&277,446&7.609&7.722&8.241&1,097&1,101&1,125\\
		    FiftyWords&279,274&281,506&292,386&6.052&6.353&6.428&1,129&1,138&1,183\\
		    Fish&278,015&281,791&300,391&3.770&3.850&3.912&1,125&1,139&1,214\\
		    FordA&278,218&282,290&302,370&43.135&44.861&47.525&1,124&1,142&1,221\\
		    FordB&278,218&282,290&302,370&26.781&27.341&27.890&1,124&1,142&1,221\\
		    GunPoint&269,818&271,090&277,170&1.003&1.046&1.138&1,092&1,098&1,123\\
		    Ham&276,562&280,082&297,402&2.048&2.127&2.160&1,118&1,133&1,202\\
		    HandOutlines&331,234&352,978&461,418&61.902&62.375&63.393&1,332&1,418&1,842\\
		    Haptics&292,837&301,645&345,405&9.787&10.023&10.631&1,183&1,217&1,390\\
		    Herring&278,506&282,674&303,234&1.633&1.668&1.706&1,126&1,144&1,225\\
		    InlineSkate&312,071&327,199&402,559&16.439&16.772&17.853&1,258&1,317&1,614\\
		    InsWingSound&273,595&275,715&286,035&4.332&4.510&4.599&1,107&1,115&1,158\\
		    ItalyPowD&266,794&267,058&268,098&2.719&3.015&3.048&1,080&1,083&1,087\\
		    LargeKApp&283,635&289,467&318,347&10.786&12.008&11.640&1,147&1,170&1,283\\
		    Lightening2&281,506&286,674&312,234&3.887&3.940&4.065&1,137&1,159&1,260\\
		    Lightening7&274,559&277,183&290,023&4.091&4.811&4.477&1,111&1,121&1,174\\
		    MALLAT&291,616&299,880&340,920&34.911&37.448&38.080&1,178&1,210&1,373\\
		    Meat&277,107&280,763&298,763&1.698&1.737&1.832&1,122&1,136&1,207\\
		    MedicalImages&269,690&270,554&274,594&5.361&5.456&6.498&1,092&1,095&1,114\\
		   MidPhOAgeGrp&268,275&268,987&272,267&1.802&2.138&2.182&1,087&1,090&1,103\\
		   MidPhOCorrect&268,138&268,850&272,130&3.219&3.374&3.528&1,085&1,090&1,103\\
		   MidPhTW&268,686&269,398&272,678&2.271&2.340&2.321&1,088&1,091&1,106\\
		   MoteStrain&268,234&268,978&272,418&2.398&2.423&2.481&1,085&1,090&1,104\\
		   NonInvECGTh1&289,698&295,770&325,850&61.809&61.853&71.308&1,170&1,194&1,314\\
		   NonInvECGTh2&289,698&295,770&325,850&59.212&60.554&60.754&1,170&1,194&1,314\\
		   OliveOil&280,172&284,804&307,684&3.267&3.670&4.073&1,133&1,151&1,243\\
		   OSULeaf&277,014&280,502&297,662&4.962&5.096&5.409&1,121&1,134&1,204\\
		   PhalOCorrect	&268,138&	268,850	&272,130	&	16.319&	19.269&	21.159	&	1,085&	1,090&	1,103\\
		   Phoneme&	295,863	&304,127&	345,167	&	29.778&	31.34&	37.147&		1,194&	1,226&	1,389\\
		   Plane&	270,359&	271,583&	277,423	&	0.497&	0.502&	0.575	&	1,095&	1,099&	1,125\\
		   ProxPhOAgeGrp	&268,275&	268,987	&272,267&		3.550&	3.601&	3.605&		1,087&	1,090	&1,103\\
		   ProxPhOCorrect&	268,138	&268,850&	272,130	&	4.142	&4.538	&4.678&		1,085&	1,090&	1,103\\
		   ProxPhTW	&268,686&	269,398	&272,678	&	2.050&	2.201&	2.126	&	1,088&	1,091&	1,106\\
		   RefDevices&	283,635&	289,467&	318,347	&	12.878&	14.160&	14.460&		1,147&	1,170&	1,283\\
		   ScreenType&	283,635&	289,467&	318,347	&	13.327&	13.890&	14.283	&	1,147&	1,170&	1,283\\
		   ShapeletSim	&278,218&	282,290	&302,370&		1.596&	1.628&	2.004	&	1,124	&1,142	&1,221\\
		   ShapesAll&	286,452	&290,620	&311,180	&	34.243&	36.523&	37.627	&	1,157&	1,173&	1,256\\
		   SmlKitApp&	283,635&	289,467	&318,347	&	12.417&	12.92&	14.248	&	1,147&	1,170&	1,283\\
		   SonyAIBORI&	267,898&	268,530	&271,410	&	0.982&	1.931&	2.042	&	1,084&	1,088&	1,100\\
		   SonyAIBORII &	267,778	&268,370	&271,050	&	2.492&	2.496&	2.873	&	1,084&	1,088&	1,099\\
		   StarLightCurves 	&290,931&	299,195	&340,235	&	151.538&	157.143&	161.447	&	1,176&	1,208&	1,369\\
		   Strawberry&	271,858	&273,810&	283,290	&	39.138&	40.408&	42.769	&	1,100&	1,109&	1,147\\
		   SwedishLeaf &	271,071	&272,167&	277,367	&	6.931&	7.572&	7.891	&	1,098&	1,102&	1,125\\
		   Symbols&	276,318	&279,574&	295,574	&	6.176&	6.543&	6.736&		1,118&	1,131&	1,196\\
		   SynControl &	268,206	&268,758&	271,238	&	20.562&	21.735&	23.209	&	1,086&	1,088&	1,101\\
		   ToeSeg1&	272,866	&275,154&	286,314	&	1.824&	1.846&	1.900&		1,104&	1,114&	1,158\\
		   ToeSeg2&	274,450&	277,266&	291,066	&	1.415&	1.549&	1.629&		1,110&	1,122&	1,177\\
		   Trace &	273,092&	275,364	&286,444	&	0.977&	1.021&	1.093&		1,105&	1,114&	1,160\\
		   TwoLeadECG&	268,186	&268,914&	272,274	&	3.053&	3.535&	3.498	&	1,085&	1,090&	1,104\\
		   TwoPatterns&	269,564	&270,660&	275,860	&	33.994&	37.673&	38.303	&	1,092&	1,096	&1,119\\
	   		UWaveAll&	289,720	&297,352&	335,232	&	24.983&	28.702&	28.874	&	1,170&	1,200&	1,351\\
	   		UWaveX &	274,600	&277,192&	289,872	&	30.214&	32.095&	33.573	&	1,111&	1,121&	1,173\\
	   		UWaveY	&274,600&	277,192	&289,872	&	30.214&	31.526	&32.526	&	1,111&	1,121&	1,173\\
	   		UWaveZ&	274,600	&277,192&	289,872	&	30.214&	31.881&	33.573	&	1,111&	1,121&	1,173\\
	   		Wafer &	269,866	&271,154	&277,314	&	20.438&	21.835&	22.018	&	1,092&	1,099&	1,123\\
	   		Wine&	271,834	&273,778	&283,218	&	3.771&	4.021&	4.530	&	1,099&	1,109&	1,146\\
	   		WordSynonyms &	275,849	&278,081&	288,961	&	4.911&	5.155&	5.498&		1,116&	1,125&	1,170\\
	   		Worms&	288,229	&295,501&	331,581	&	4.484&	4.669&	5.019	&	1,165&	1,193&	1,336\\
	   		WormsTwoClass&	287,818&	295,090	&331,170	&	3.536&	3.586&	4.134	&	1,162&	1,192&	1,334\\
	   		Yoga &	276,442	&279,922&	297,042	&	10.970&	11.606&	10.753	&	1,118&	1,133	&1,200\\
	   		\Xhline{2\arrayrulewidth}
		    Total &\textbf{23,555,876}&	23,849,100&	25,291,420&
		    \textbf{1145.645}&	1207.916 &	1251.71&\textbf{95,273}&96,480&100,992\\
			\hline
			\label{parameters_compare}
		\end{tabular}
	\end{center}
\end{table*}

\begin{table} 
	\begin{center}
		\renewcommand{\arraystretch}{0.9}
		\scriptsize
		\caption{The f1-score value of the proposed GRU-FCN model and the LSTM-based architectures over the UCR benchmark datasets~\cite{ucr_benchmark}.}
		\begin{tabular}{l| l l l }%{p{2.2cm} p{1.5cm}p{1cm}p{1cm}p{1cm}p{1cm}p{1cm}p{1cm}p{0.7cm}p{0.7cm}}
			\hline
			\textbf{Dataset}&\multicolumn{3}{|c}{\textbf{f1-Score}}\\
			\cline{2-4} 
			%	\multicolumn{}{| l |}
			
			\textbf{} &\textbf{\textit{GRU-FCN}}& \textbf{\textit{LSTM-FCN}}& \textbf{\textit{ALSTM-FCN}}\\
			\hline
			Adiac&	\textbf{0.795}&	0.770&	0.780\\
			ArrowHead&	\textbf{0.711}&	0.694&	0.695\\
			Beef&	0.819&\textbf{0.873}&	0.765\\
			BeetleFly&	\textbf{1.0}&\textbf{1.0}&	0.949\\
			BirdChicken	&\textbf{1.0}&\textbf{1.0}&\textbf{1.0}\\
			Car	&\textbf{0.954}&0.952&	0.947\\
			CBF&\textbf{0.995}&	0.994&	0.989\\
			ChlorineCon&0.766&\textbf{0.791}&0.767\\
			CinCECGTorso &\textbf{0.379}&0.321&0.375\\
			Coffee &\textbf{1.0}&\textbf{1.0}&\textbf{1.0}\\
			Computers&\textbf{0.916}&	0.914&	0.913\\
			CricketX&\textbf{0.786}&0.782&0.784\\
			CricketY&0.756&\textbf{0.786}&0.776\\
			CricketZ &	\textbf{0.779}&	0.778&	0.761\\
			DiatomSizeR &	0.926&	0.926&	\textbf{0.935}\\
			DisPhOAgeGrp&\textbf{0.645}&	0.614&	0.636\\
			DisPhOCorrect&	\textbf{0.813}&	0.804&\textbf{0.813}\\
			DisPhTW&	0.477&	0.469&\textbf{0.479}\\
			Earthquakes	&\textbf{0.483}&	0.466&	0.466\\
			ECG200	&\textbf{0.910}&	0.900&	0.909\\
			ECG5000&	0.253&	0.251&	\textbf{0.263}\\
			ECGFiveDays &	\textbf{0.991}&	\textbf{0.991}	&\textbf{0.991}\\
			ElectricDevices&	0.195&	0.196&	\textbf{0.197}\\
			FaceAll &	\textbf{0.137}&	0.134&	0.136\\
			FaceFour &\textbf{0.960}&	0.949&	0.949\\
			FacesUCR &	0.892&	\textbf{0.898}&	0.896\\
			50words &	\textbf{0.353}&	0.330&	\textbf{0.353}\\
			Fish& 	0.962&\textbf{0.964}&	0.957\\
			FordA&	0.926&	\textbf{0.928}&	\textbf{0.928}\\
			FordB&	0.928&	\textbf{0.930}&	0.929\\
			GunPoint &	\textbf{1.0}&\textbf{1.0}&\textbf{1.0}\\
			Ham&	\textbf{0.788}&	\textbf{0.788}&	0.770\\
			HandOutlines&	\textbf{0.875}&	0.873&	0.866\\
			Haptics &\textbf{0.528}&	0.523&	0.515\\
			Herring&	0.717&	\textbf{0.722}&	0.694\\
			InlineSkate	&0.454	&\textbf{0.474}&	0.446\\
			InWingSound	&\textbf{0.477}&	0.432&	0.410\\
			ItalyPower 	&0.970&	0.970&	\textbf{0.972}\\
			LargeKApp&	0.406&	0.407&	\textbf{0.410}\\
			Lightning2 &	0.765&	\textbf{0.767}&	\textbf{0.767}\\
			Lightning7 	&\textbf{0.872}&	0.833&	0.858\\
			MALLAT&	0.971&	0.970&	0.971\\
			Meat&	0.925&	0.870&	\textbf{0.973}\\
			MedicalImages&	\textbf{0.714}&	0.686&	0.701\\
			MidPhOutlineAgeGrp&	\textbf{0.507}&	0.347&	0.445\\
			MidPhOCorrect&\textbf{0.823}&	0.821&	0.819\\
			MidPhTW&	\textbf{0.329}&	0.314&	0.320\\
			MoteStrain	&0.925&	0.920&	0.915\\
			NonInvECGTh1 &	\textbf{0.911}&	0.908&	0.905\\
			NonInvECGTh2 &	\textbf{0.899}&	0.896&	0.894\\
			OliveOil&	0.853&	0.611&\textbf{0.885}\\
			OSULeaf	&\textbf{0.988}&	0.979&	\textbf{0.988}\\
			PhalOCorrect&	\textbf{0.812}&	0.803&	0.809\\
			Phoneme	&0.025&	\textbf{0.026}&	\textbf{0.026}\\
			Plane&	\textbf{0.888}&	\textbf{0.888}&	0.882\\
			ProxPhOeAgeGrp&	\textbf{0.600}&	0.594&	0.436\\
			ProxPhOCorrect&	0.896&	\textbf{0.904}&	0.896\\
			ProxPhTW&	0.545	&0.504&	0.469\\
			RefDevices&	\textbf{0.277}&	0.241&	0.241\\
			ScreenType&	0.297&	0.302&	\textbf{0.308}\\
			ShapeletSim	&\textbf{0.842}&	\textbf{0.842}&	\textbf{0.842}\\
			ShapesAll&\textbf{0.108}&	\textbf{0.108}&	0.107\\
			SmlKitApp&	0.345&	0.361&	\textbf{0.370}\\
			SonyAIBORI&	\textbf{0.984}&	0.974&	0.983\\
			SonyAIBORII &	\textbf{0.980}&	0.978&	0.977\\
			StarLightCurves &	\textbf{0.975}&	0.961&	0.962\\
			Strawberry&	\textbf{0.818}&	\textbf{0.818}&	\textbf{0.818}\\
			SwedishLeaf &	0.807&	0.801&	\textbf{0.811}\\
			Symbols&	0.980&	\textbf{0.982}&	0.974\\
			SynControl &	\textbf{0.522}&	0.516&	0.511\\
			ToeSeg1	&0.708	&\textbf{0.746}&	\textbf{0.746}\\
			ToeSeg2	&\textbf{0.582}&	0.563&	0.577\\
			Trace &\textbf{1.0}&	0.986&	0.983\\
			TwoLeadECG&	\textbf{0.999}&	\textbf{0.999}	&\textbf{0.999}\\
			TwoPatterns	&0.986&	\textbf{0.989}	&0.971\\
			UWaveAll&	\textbf{0.782}&	0.766&	0.754\\
			UWaveX 	&\textbf{0.665}&	0.654&	0.659\\
			UWaveY&	\textbf{0.698}&	0.695&	0.686\\
			UWaveZ&	0.736&	\textbf{0.739}&	0.743\\
			Wafer &	\textbf{0.996}&	\textbf{0.996}&	\textbf{0.996}\\
			Wine&\textbf{0.887}&	\textbf{0.887}&	\textbf{0.887}\\
			WordSynonyms &	\textbf{0.380}&	0.327&	0.345\\
			Worms&	\textbf{0.448}&	0.423&	0.425\\
			WormsTwoClass&	0.530&	0.525&	\textbf{0.542}\\
			Yoga &	0.882&	0.906&	\textbf{0.914}\\		
			\hline
		\end{tabular}
		\label{f1_score_table}
	\end{center}
\end{table}

% model with error rate $e_m$ when applied to a specific dataset $m$ consisting of $c_m$ class labels, and $M$ is the total number of datasets that are applied.
\begin{table*}%[h]
	\caption{Wilcoxon signed-rank test on GRU-FCN and 10 benchmark model on the 85 datasets from UCR benchmark~\cite{ucr_benchmark}.}
	\setlength{\tabcolsep}{2.5pt}
	\begin{center}
		\begin{tabular}{ l l l l l l l l l l l l l}
			\hline
			\textbf{ } &\textbf{\textit{FCN}}& \textbf{\textit{LSTM-FCN}}&\textbf{\textit{ALSTM-FCN}}&\textbf{\textit{ResNet}}&\textbf{\textit{MCNN}}&\textbf{\textit{MLP}}&\textbf{\textit{COTE}}&\textbf{\textit{DTW}}&\textbf{\textit{PROP}}&\textbf{\textit{BOSS}}&\textbf{\textit{TSBF}}&\textbf{\textit{ED}}\\
			\hline
			\textbf{GRU-FCN}&\textbf{3.44E-10}&\textbf{4.95E-03}&\textbf{4.00E-02}&2.53E-11&1.05E-12&1.43E-13&1.25E-08&1.23E-14&1.58E-11&4.37E-10&2.77E-12&2.93E-15\\
			\textbf{FCN}&&4.37E-09&8.58E-08&\textbf{1.68E-01}&9.31E-10&1.12E-09&\textbf{1.85E-02}&3.49E-12&1.31E-07&8.02E-04&1.10E-07&7.07E-13\\
			\textbf{LSTM-FCN}&&&\textbf{7.45E-01}&2.24E-09&1.40E-11&6.09E-13&1.03E-06&2.35E-14&5.72E-11&2.85E-9&6.40E-13&1.08E-14\\			
			\textbf{ALSTM-FCN}&&&&\textbf{1.40E-07}&1.02E-11&8.35E-12&\textbf{2.33E-07}&1.55E-13&7.95E-11&4.71E-09&3.30E-12&4.73E-14\\		
			\textbf{ResNet}&&&&&\textbf{6.28E-09}&1.79E-08&\textbf{2.46E-01}&9.32E-13&1.76E-06&\textbf{1.28E-03}&1.56E-07&1.11E-13\\		
			\textbf{MCNN}&&&&&&\textbf{4.35E-05}&4.77E-08&5.76E-05&\textbf{6.10E-04}&1.20E-06&7.72E-06&\textbf{2.18E-04}\\
			\textbf{MLP}&&&&&&&7.04E-05&\textbf{7.28E-01}&\textbf{7.13E-01}&\textbf{1.08E-03}&5.70E-03&3.25E-04\\			
			\textbf{COTE} &&&&&&&&1.62E-06&2.28E-05&\textbf{7.74E-03}&3.59E-04&3.22E-07\\
			\textbf{DTW}&&&&&&&&&\textbf{2.05E-01}&2.37E-07&1.80E-04&2.13E-03\\			
			\textbf{PROP}&&&&&&&&&&\textbf{8.82E-03}&\textbf{5.13E-01}&\textbf{3.14E-02}\\
			\textbf{BOSS} &&&&&&&&&&&\textbf{3.18E-02} &7.02E-10\\
			\textbf{TSBF} &&&&&&&&&&&&\textbf{6.65E-08}\\
			\hline
		\end{tabular}
		\label{wilcoxon_signed_rank_test}
	\end{center}
\end{table*}

\begin{figure*}%[h]
	\centering
	\includegraphics[width=\textwidth,height=4.5cm]{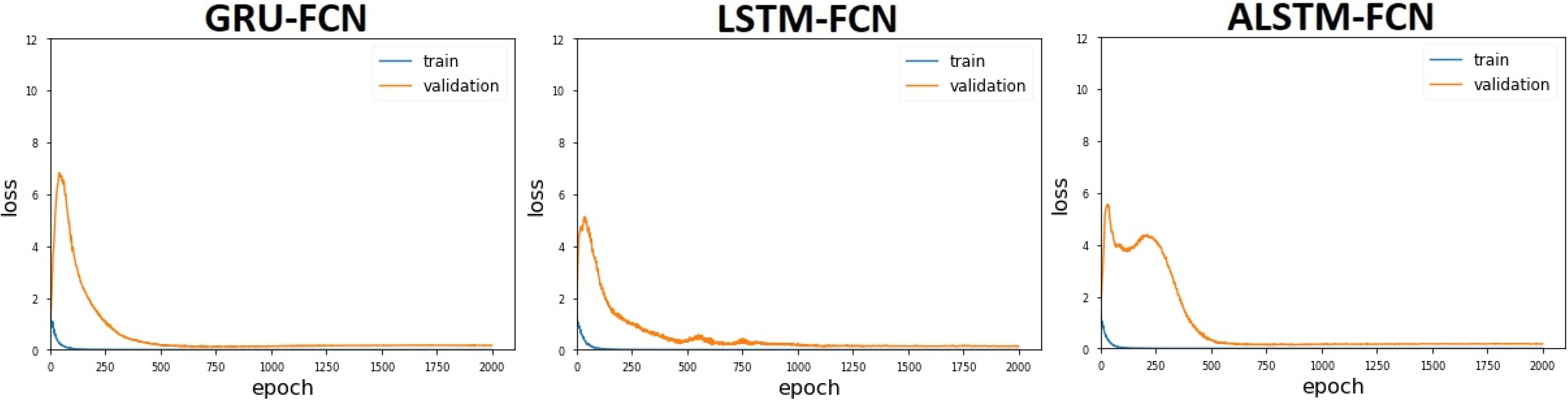}
	\caption{The loss value of GRU-FCN, LSTM-FCN, and ALSTM-FCN models over the image-source obtained (DiatomSizeR dataset) training and validation processes.}
	\label{diatatom}
\end{figure*}

\begin{figure*}%[h]
	\centering
	\includegraphics[width=\textwidth,height=4.5cm]{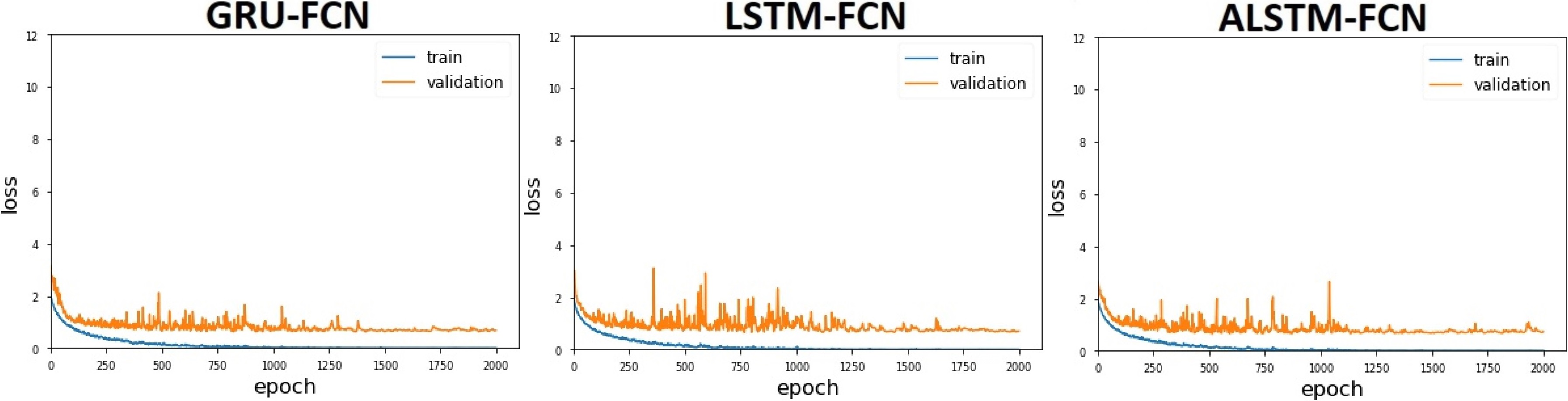}
	\caption{The loss value of GRU-FCN, LSTM-FCN, and ALSTM-FCN models over the motion-source obtained (CricketX dataset) training and validation processes.}
	\label{cricketX_diagram}
\end{figure*}

\begin{figure*}%[h]
	\centering
	\includegraphics[width=\textwidth,height=4.5cm]{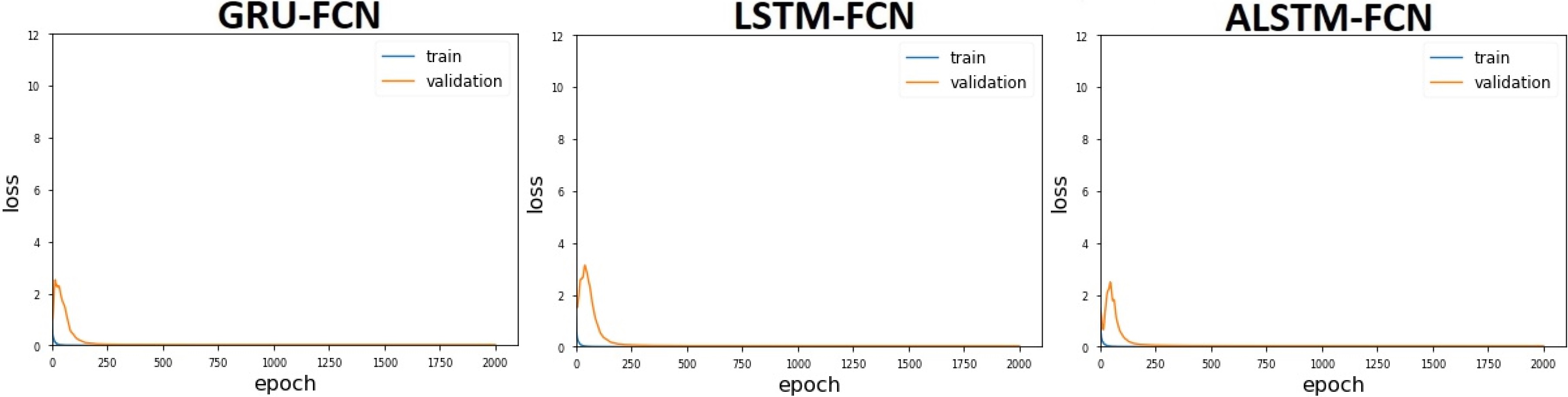}
	\caption{The loss value of GRU-FCN, LSTM-FCN, and ALSTM-FCN models over the simulated-source obtained (CDF dataset) training and validation processes.}
	\label{cbf_diagram}
\end{figure*}

\begin{figure*}%[h]
	\centering
	\includegraphics[width=\textwidth,height=4.5cm]{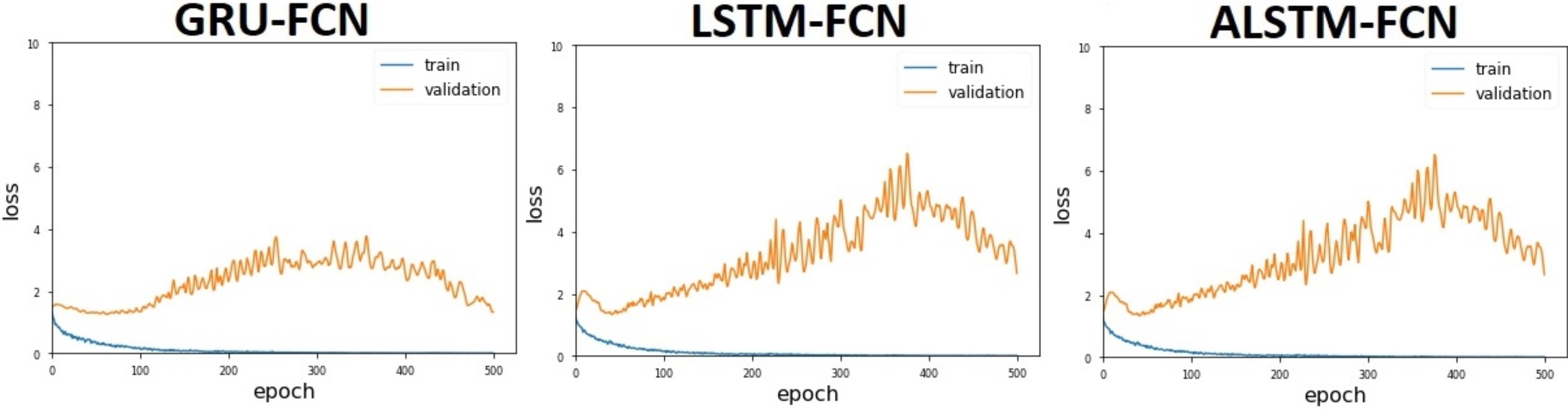}
	\caption{The loss value of GRU-FCN, LSTM-FCN, and ALSTM-FCN models over the sensor-source obtained (ChlorineCon dataset) training and validation processes.}
	\label{chlorine_diagram}
\end{figure*}

Table~\ref{table1} shows the MPCE value for our GRU-FCN and other state-of-the-art models on the UCR benchmark datasets~\cite{ucr_benchmark}. 
The results obtained by implementing GRU-FCN and generating LSTM-FCN, and ALSTM models based on their actual implementation on github. 
For the other models, we obtained the results from their own publications. Our GRU-FCN has the smallest MPCE value compared to the other state-of-the-art classification models. This means that generally our GRU-FCN model performance across the different datasets is higher than the other state-of-the-art models. 

Figures~\ref{diatatom},~\ref{cricketX_diagram},~\ref{cbf_diagram},~\ref{chlorine_diagram}  are showing the loss value of both the training and validation processed of datasets. Each of these figures represents the loss process over image, motion, simulated, and source-obtained datasets from the UCR benchmark datasets respectively. These figures show that the average difference between the training and validation loss for the GRU-FCN is smaller that the LSTM-FCN and ALSTM-FCN models.

Table~\ref{f1_score_table} shows the f1-score (also known as F-score or F-measure)~\cite{sasaki2007truth,powers2011evaluation} for GRU-FCN, LSTM-FCN, and ALSTM-FCN classifiers. The f1-score shows the overall measure of a model\textquotesingle s accuracy over each dataset used. The f1-score measuring based on both the precision and recall values of the classification model~\cite{sasaki2007truth,powers2011evaluation}. The f1-score is calculated as follows~\cite{sasaki2007truth,powers2011evaluation}:

\begin{flalign}
%\begin{equation}
precision &= \frac{TP}{TP+FP}\\
%\end{equation}
%\begin{equation}
recall&= \frac{TP}{TP+FN}\\
%%\end{equation}
%\begin{equation}
\mathit{f1\textnormal{-}score} &= 2 \times \frac{precision \times recall}{precision + recall}
%\end{equation}
\end{flalign}
\noindent
where TP, FP, FN stands for true-positive, false-positive and false-negative respectively. The GRU-FCN shows the highest f1-score for $\mathrm{53}$ out of $\mathrm{85}$ datasets comparing to the LSTM-FCN and ALSTM-FCN that both of these models have the highest f1-score for only $\mathrm{29}$ out of $\mathrm{85}$ datasets.

Figure~\ref{critical_diagram} shows the critical difference diagram \cite{criticalDiagram} for Nemenyi or Bonferroni-Dunn test~\cite{critical_test} with $\alpha=0.05$ on our GRU-FCN and the state-of-the-art models based on the ranks arithmetic mean on the UCR benchmark datasets. This graph shows the significant classification accuracy improvement of our GRU-FCN compared to the other state-of-the-art models.

The Wilcoxon signed-rank test is one of substantial tests to provide the classification method efficiency~\cite{woolson2007wilcoxon,rey2011wilcoxon}. Table~\ref{wilcoxon_signed_rank_test} shows the Wilcoxon signed-rank test~\cite{lowry2014concepts,woolson2007wilcoxon} among the twelve state-of-the-art classification models. This provides the overall accuracy evidence of each of the twelve classification methods.

\section{Conclusion}
The proposed GRU-FCN classification model shows that replacing the LSTM by a GRU
%in classification problems 
enhances the classification accuracy without needing extra algorithm enhancements such as fine-tuning or attention algorithms. 
The GRU also has a smaller architecture that requires fewer computations than the LSTM. Moreover, the GRU-based model requires smaller number of trainable parameters, memory, and training time comparing to the LSTM-based models.
%Moreover, the model does not require any data preprocessing or feature engineering prior to the classification training and testing stages. 
Furthermore, the proposed GRU-FCN classification model achieves the performance of state-of-the-art models and has the highest average arithmetic ranking and the lowest mean per-class error (MPCE) through time series datasets classification of the UCR benchmark compared to the state-of-the-art models. Therefore, replacing the LSTM by GRU in the LSTM-FCN for univariate time series classification can improve the classification with smaller model architecture.

\section{Acknowledgment}
We would like to thank Prof. Eamonn Keogh for his suggestions, comments, and ideas about our proposed work in this paper.

\bibliographystyle{IEEEtran}
\bibliography{timeSeriesReferences}
\begin{IEEEbiography}[{\includegraphics[width=1.5in,height=1.25in,clip,keepaspectratio]{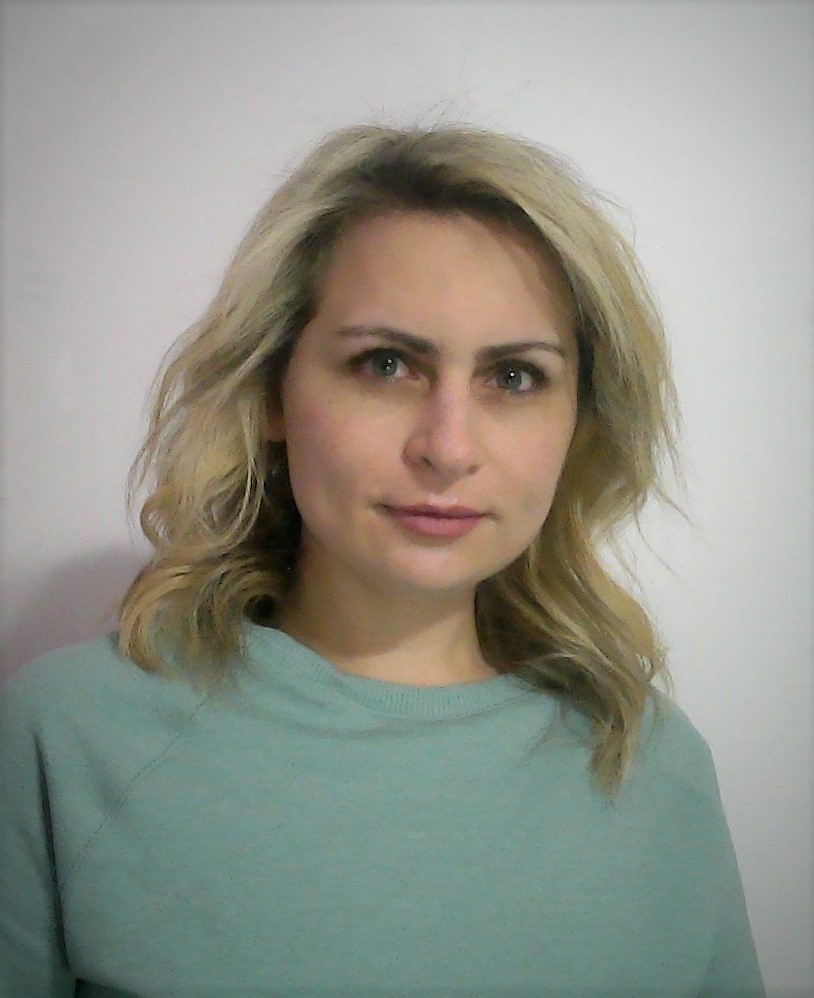}}]{Nelly Elsayed}
	is a PhD candidate in Computer Engineering at the School of Computing and Informatics, University of Louisiana at Lafayette. She received her Bachelor degree in Computer Science at Alexandria University in 2010 and she was ranked the First on Computer Science Class. She received two Masters degrees, the first in Computer Science in 2014 at Alexandria University and the second in Computer Engineering in 2017 at University of Louisiana at Lafayette. She is an honor member in the national society of leadership and the the honor society Phi Kappa Phi. Her major research interests are in machine learning, deep learning, artificial intelligence, convolutional recurrent neural networks, bio-inspired computations, and quantum computing. 
\end{IEEEbiography}
\begin{IEEEbiography}[{\includegraphics[width=1.5in,height=1.25in,clip,keepaspectratio]{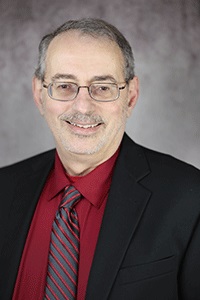}}]{Anthony S. Maida}
	is associate professor and graduate coordinator for computer science and computer engineering at School of Computing and Informatics, University of Louisiana at Lafayette. He received this BA in Mathematics in 1973, Ph.D. in Psychology in 1980, and Masters Degree in Computer Science in 1981, all from the University of Buffalo. He has done two Post Doctoral degrees at Brown University and the University of California, Berkeley. He was a member of the computer science faculty at the Penn State University from 1984 through 1991. He has been a member of a Center for Advanced Computer Studies and School of Computing and informatics at the University of Louisiana at Lafayette from 1991 to the present. His research interests are: intelligent systems, neural networks, recurrent neural networks and brain simulation.
\end{IEEEbiography}
\begin{IEEEbiography}[{\includegraphics[width=1.5in,height=1.25in,clip,keepaspectratio]{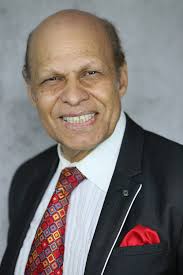}}]{Magdy Bayoumi}
	is department head and professor at the Electrical Engineering Department, University of Louisiana at Lafayette. He was the Director of CACS, 1997 – 2013 and Department Head of the Computer Science Department, 2000-2011. He has been a faculty member in CACS since 1985. He received B.Sc. and M.Sc. degrees in Electrical Engineering from Cairo University, Egypt; M.Sc. degree in Computer Engineering from Washington University, St. Louis; and Ph.D. degree in Electrical Engineering from the University of Windsor, Canada. He is on the IEEE Fellow Committee and he was on the IEEE CS Fellow Committee. His research interests are: technology, data processing, management, energy and security.
	
\end{IEEEbiography}
%\EOD

\end{document}